\documentclass[10pt,twocolumn,letterpaper]{article}
\usepackage{cvpr}
\usepackage{times}
\usepackage{epsfig}
\usepackage{graphicx}
\usepackage{amsmath}
\usepackage{amssymb}
\usepackage{subfigure}
\usepackage[pagebackref=true,breaklinks=true,letterpaper=true,colorlinks,bookmarks=false]{hyperref}
\cvprfinalcopy 

\begin{document}
\title{Structure Inference Net: Object Detection Using Scene-Level Context and Instance-Level Relationships}

\author{Yong Liu$^{1,2}$, Ruiping Wang$^{1,2,3}$, Shiguang Shan$^{1,2,3}$, Xilin Chen$^{1,2,3}$\\
$^1$Key Laboratory of Intelligent Information Processing of Chinese Academy of Sciences (CAS),\\
Institute of Computing Technology, CAS, Beijing, 100190, China\\
$^2$University of Chinese Academy of Sciences, Beijing, 100049, China\\
$^3$Cooperative Medianet Innovation Center, China\\
{\tt\small yong.liu@vipl.ict.ac.cn, \{wangruiping, sgshan, xlchen\}@ict.ac.cn}}

\maketitle
\thispagestyle{empty}

\begin{abstract}
Context is important for accurate visual recognition. In this work we propose an object detection algorithm that not only considers object visual appearance, but also makes use of two kinds of context including scene contextual information and object relationships within a single image. Therefore, object detection is regarded as both a cognition problem and a reasoning problem when leveraging these structured information. Specifically, this paper formulates object detection as a problem of graph structure inference, where given an image the objects are treated as nodes in a graph and relationships between the objects are modeled as edges in such graph. To this end, we present a so-called Structure Inference Network (SIN), a detector that incorporates into a typical detection framework (e.g. Faster R-CNN) with a graphical model which aims to infer object state. Comprehensive experiments on PASCAL VOC and MS COCO datasets indicate that scene context and object relationships truly improve the performance of object detection with more desirable and reasonable outputs.
\end{abstract}

\vspace{-2ex}
\section{Introduction}

Object detection is one of the fundamental computer vision problems. Recently, this topic has enjoyed a series of breakthroughs thanks to the advances of deep learning, and it is observed that prevalent object detectors predominantly regard detection as a problem of classifying candidate boxes \cite{RCNN, Fast, Faster, FPN, RFCN}. While most of them have achieved impressive performance in a number of detection benchmarks, they only focus on local information near an object's region of interest within the image. Usually an image contains rich contextual information including scene context and object relationships \cite{AES}. Ignoring these information inevitably places constraints on the accuracy of objects detected \cite{ION}.

\begin{figure}[t]
\vspace{-1.5ex}
\hspace{-1ex}
\centering
\subfigure[] {
 \label{O}
 \includegraphics[width=0.48\linewidth]{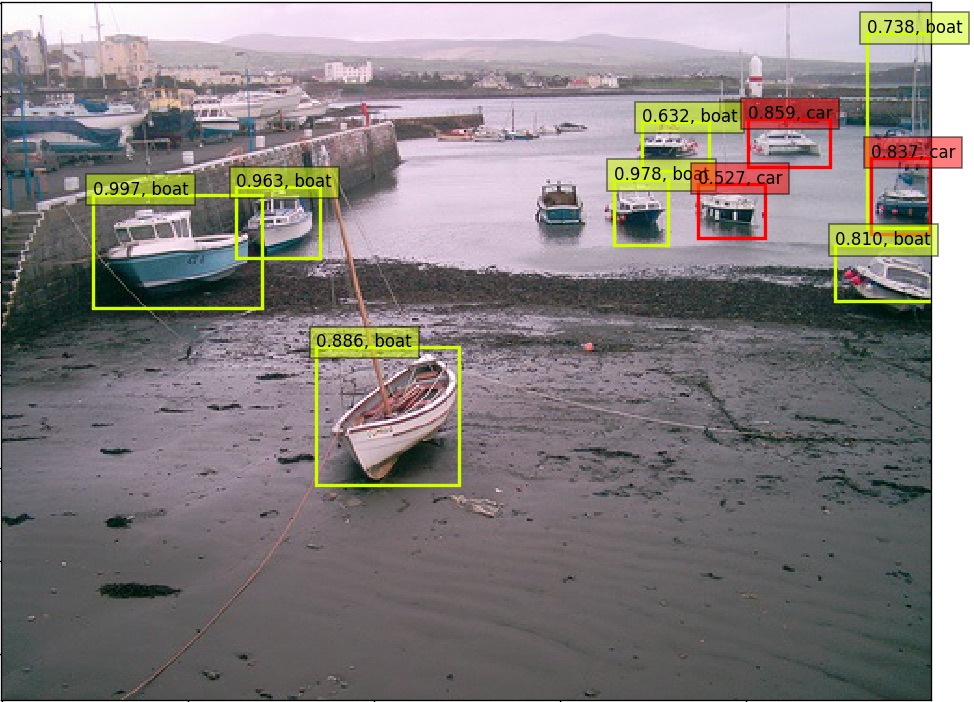}} 
 \hspace{-1ex}
 \subfigure[] {
 \label{O}
 \includegraphics[width=0.46\linewidth]{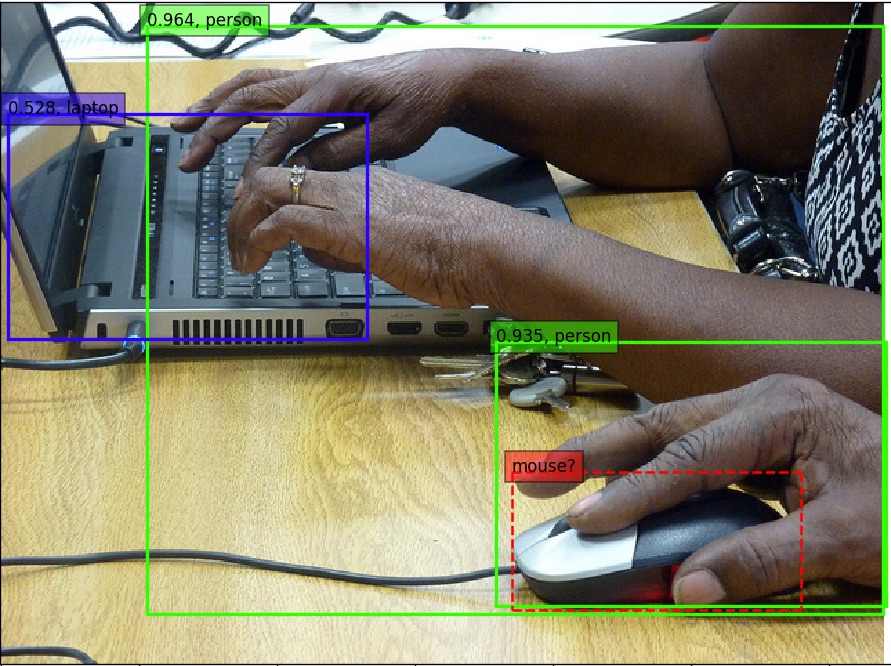}}
\caption{{\bf Some Typical Detection Errors of Faster R-CNN.} (a) Some {\em boats} are mislabeled as {\em cars} on PASCAL VOC \cite{voc}. (b) The {\em mouse} is undetected on MS COCO \cite{coco}.}
\label{fig:errors}
\vspace{-2.5ex}
\end{figure}

To illustrate such constraints, considering the practical examples in Fig. \ref{fig:errors}, detected by Faster R-CNN \cite{Faster}. In the first case where is a river field, some of the {\em boats} are mislabeled as {\em cars}, since the detector only concentrates on object's visual appearance. If the scene information in this image was taken into account, such banana skin could have been easily avoided. In the second case, though a {\em laptop} and {\em person} have been detected as expected, no further object is found any more. It is quite common that mouse and laptop usually co-occur within a single image. If using object relative position and co-occurrence pattern, more objects within the given image could be detected.

Many empirical studies \cite{AES, COLA, LSC, CBVS, Context, TRC, TB} have suggested that recognition algorithms can be improved by proper modeling of context. To handle the problem above, two types of contextual information model have been explored for detection \cite{SM}. The first type incorporates context around object or scene-level context \cite{ION, GBD, CPF}, and the second models object-object relationships at instance-level \cite{EPC, SM, Context}. While these two types of models capture complementary contextual information, they can be combined together to jointly help detection.

\begin{figure}[t]
\begin{center}
   \includegraphics[width=0.9\linewidth]{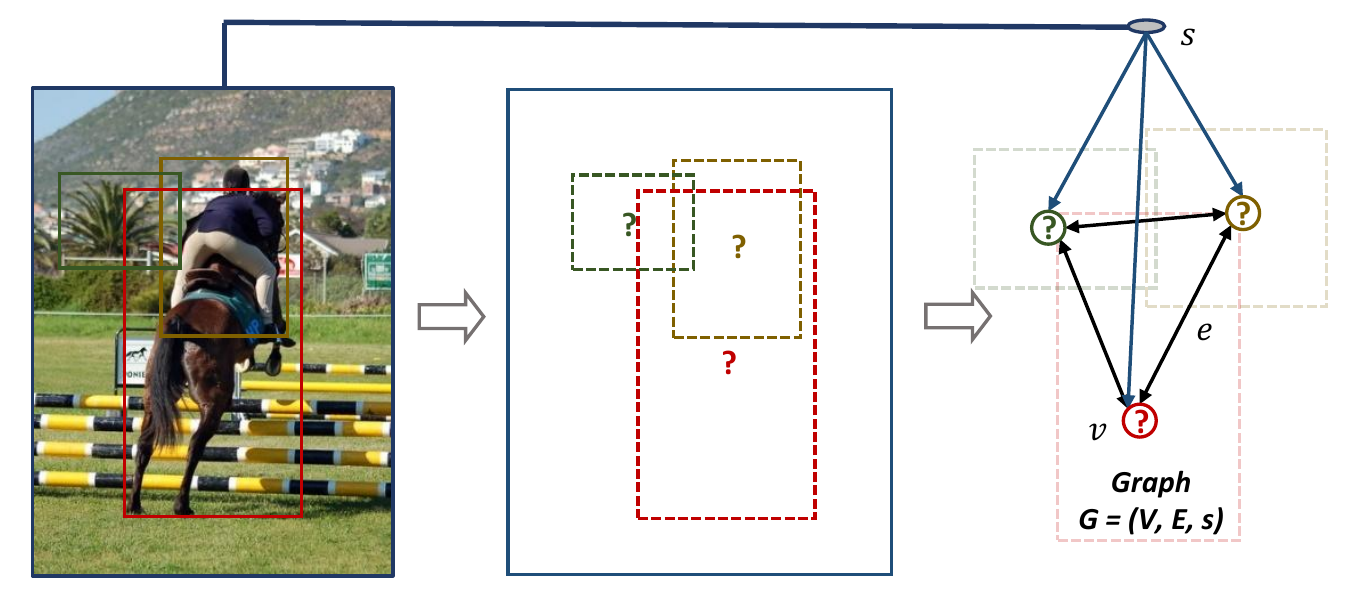}
\end{center}
\vspace{-2ex}
   \caption{{\bf Graph Problem.} Detection basically aims to answer: what is where. From a structure perspective, it can be formulated as a  reasoning problem of a graph involving the mutually complementary information of scene, objects and relationships.}
\label{fig:Problem}
\vspace{-2.5ex}
\end{figure}

We are thus motivated to intuitively conjecture that visual concepts in most of natural images form an organism with the key components of scene, objects and relationships, and different objects in the scene are organized in a structured manner, {\em e.g.} boats are on the river, mouse is near laptop. Sequentially object detection is regarded as not only a cognition problem, but also an inference problem which is based on contextual information with object fine-grained details. To systematically solve it, a tailored graph is formulated for each individual image. As described in Fig.  \ref{fig:Problem}, objects are nodes of the graph, and object relationships are edges of the graph. These objects interact with each other via the graph under the guidance of scene context. More specifically, an object will receive messages from the scene and other objects that are highly correlated with it. In such a way, object state is not only determined by its fine-grained appearance details but also effected by scene context and object relationship. Eventually the state of each object is used to determine its category and refine its location.

To make the above conjecture computationally feasible, we propose a structure inference network (SIN) to reason object state in a graph, where memory cell is the key module to encode different kinds of messages ({\em e.g.} from scene and other objects) into object state, and a novel way of using Gated Recurrent Units (GRUs) \cite{GRU} as the memory cell is presented in this work. Specifically, we fix object representation as the initial state of GRU and then input each kind of message to achieve the goal of updating object state. {Since SIN can accomplish inference as long as the inputs to it covers the representations of object, scene-level context and instance-level relationship, our structure inference method is not constrained to specific detection framework.}




\section{Related Work}
{\bf Object detection.} Modern CNN based object detection methods can be divided into two groups \cite{Focal, DSOD}: (i) region proposals based methods (two-stage detectors) and (ii) proposal-free methods (one-stage detectors).

With the resurgence of deep learning, two-stage detectors quickly come to dominate object detection during the past few years. Representative methods include R-CNN \cite{RCNN}, Fast R-CNN \cite{Fast}, Faster R-CNN \cite{Faster} and so on. The first stage produces numbers of candidate boxes, and then the second stage classifies these boxes into foreground classes or background. 
R-CNN \cite{RCNN} extracts CNN features from the candidate regions and applies linear SVMs as the classifier. 
To obtain higher speed, Fast R-CNN \cite{Fast} proposes a novel ROI-pooling operation to extract feature vectors for each candidate box from shared convolutional feature map. Faster R-CNN \cite{Faster} integrates proposal generation with the second-stage classifier into a single convolution network. 
More recently, one-stage detectors like SSD \cite{SSD} and YOLO \cite{YOLO} have been proposed for real-time detection with satisfactory accuracy. Anyway, detecting different objects in an image is always considered as some isolated tasks among these state-of-the-art methods especially in two-stage detectors. While such methods work well for salient objects most of the time, they are hard to handle small objects by using vague feature associated only with the object itself. 

{\bf Contextual information.} Consequently, it is natural to use richer contextual information.
In early years, a number of approaches have explored contextual information to improve object detection \cite{TRC, LSC, SODC, AES, FST, EHC, CBVS}. For example,
Mottaghi {\em et al.} \cite{TRC} propose a deformable part-based model, which exploits both local context around each candidate detection and global context at the level of the scene.
The presence of objects in irrelevant scenes is penalized in \cite{CBVS}. 
Recently, some works \cite{ION, GBD, CPF} based on deep ConvNet have made some attempts to incorporate contextual information to object detection. Contextual information outside the region of interest is integrated using spatial recurrent neural network in ION \cite{ION}. GBD-Net \cite{GBD} proposes a novel gated bi-directional CNN to pass message between features of different support regions around objects. Shrivastava {\em et al.} \cite{CPF} use segmentation to provide top-down context to guide region proposal generation and object detection.
While context around object or scene-level context has been addressed in such works \cite{ION, GBD, CPF} under the deep learning-based pipeline, they make less progress in exploring object-object relationships. On the contrary, a much recent work \cite{SM} proposes a new sequential reasoning architecture that mainly exploits object-object relationships to sequentially detect objects in an image, however, with only implicit yet weak consideration of scene-level context. Different from these existing works, our proposed structure inference network has the capability of jointly modeling both scene-level context and object-object relationships and inferring different object instances within an image from a structural and global perspective.

%

\begin{figure*}
\vspace{-3ex}
\begin{center}
	\includegraphics[width=0.95\linewidth, trim=0 100 0 80, clip]{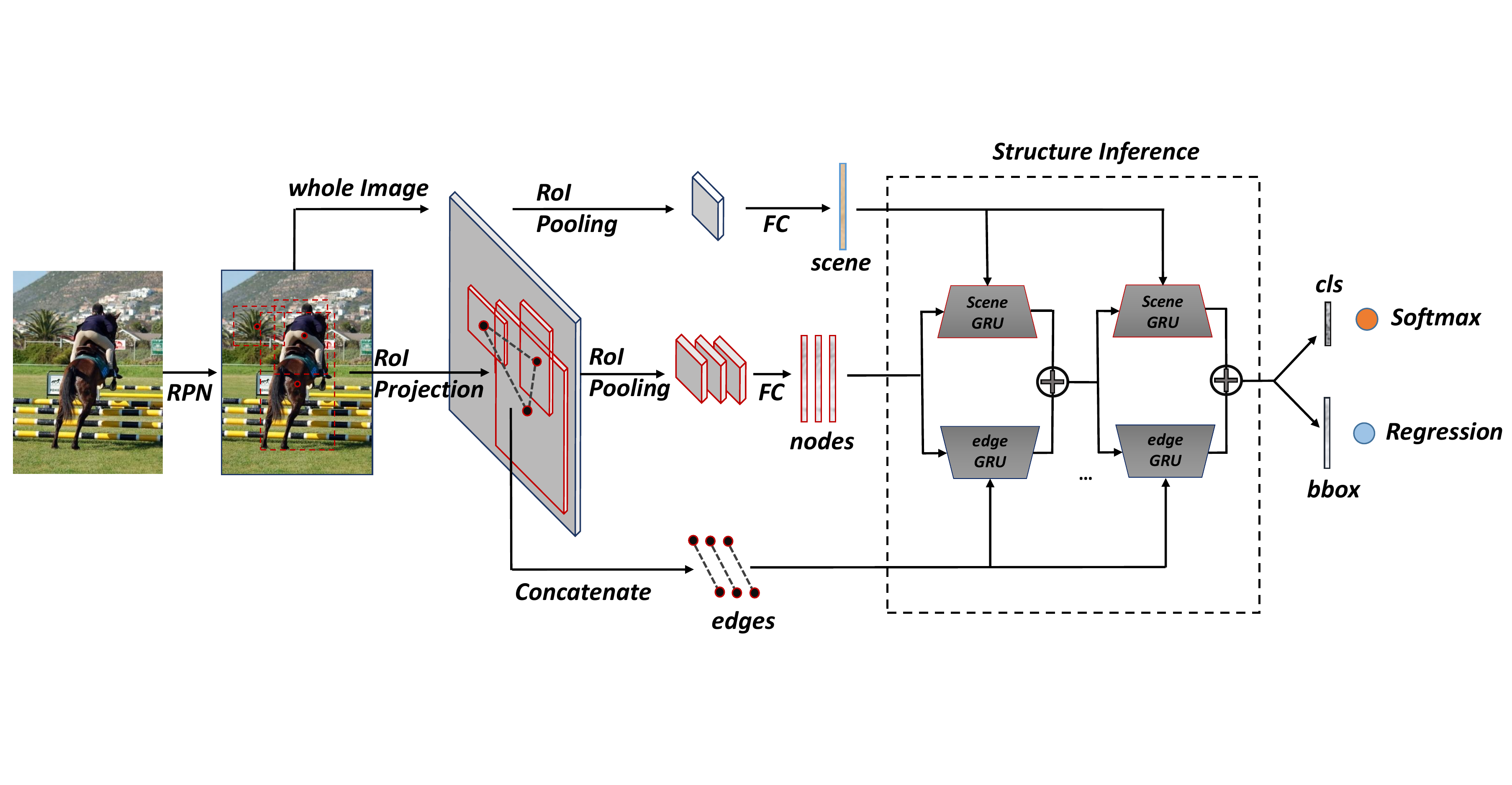}
\end{center}
\vspace{-2ex}
\caption{{\bf SIN: The Framework of Our Method.} Firstly we get a fixed number of ROIs from an input image. Each ROI is pooled into a fixed-size feature map and then mapped to a feature vector by a fully connected layer as {\em node}. We extract the whole image feature as {\em scene} in the same way, and then we concatenate the descriptors of every two ROIs into {\em edges}. To iteratively update the node state, an elaborately designed structure inference method is triggered, and the final state of each node is used to predict the category and refine the location of the corresponding ROI. The whole framework is trained end-to-end with the original multi-task loss (this study exploits Faster R-CNN as the  base detection framework).}
\label{fig:method}
\vspace{-2.5ex}
\end{figure*}

{\bf Structure inference.} Several interesting works \cite{More, GNN, GSNN, VQA, SI, SIM, IN, SRNN, SGG} have been proposed to combine deep networks with graphical models for structured prediction tasks that are solved by structure inference techniques. A generic structured model is designed to leverage diverse label relations including scene, object and attributes to improve image classification performance in \cite{SI}. Deng {\em et al.} \cite{SIM} propose structure inference machines for analyzing relations in group activity recognition. Structural-RNN \cite{SRNN} combines the power of high-level spatio-temporal graphs and sequence learning, and evaluates the model ranging from motion to object interactions. In \cite{SGG}, a graph inference model is proposed to tackle the task of generating structured scene graph from an image. 
While our work shares similar spirit as \cite{SGG} to formulate the object detection task as a graph structure inference problem, the two works have essential differences in their technical sides, such as the graph instantiation manners, inference mechanisms, message passing schemes, {\em etc}, which highly depend on the specific task domains.

\section{Method}

Our goal is to improve the detection models by exploring rich contextual information. To this end, different from existing methods that only make use of visual appearance clues, our model is designed to explicitly take object-object relationships and scene information into consideration. Specifically, a structure inference network is devised to iteratively propagate information among different objects as well as the whole scene. The whole framework of our method is depicted in Fig. \ref{fig:method}, which will be detailed in the following sections.

\subsection{Graphical Modeling}
Our structure inference network (SIN) is agnostic to the choice of base object detection framework. In this work we build SIN based on Faster R-CNN as a demonstration, which is an advanced method for detection. We present a graph $G=(V,E,s)$ to model the graphical problem as shown in Fig. \ref{fig:Problem}. The nodes $v \in V$ represent the region proposals, while $s$ is the scene of the image, and $e \in E$ is the edge (relationship) between each pair of object nodes.

Specifically, after Region Proposal Network (RPN \cite{Faster}), thousands of region proposals that might contain objects are obtained. We then use Non-Maximum Suppression (NMS \cite{NMS}) to choose a fixed number of ROIs (Region of Interest). For each ROI $v_i$, we extract the visual feature $f^{v}_{i}$ by an FC layer that follows an ROI pooling layer. For scene $s$ about the image, since there is no ground-truth scene label for the image, the whole image visual feature $f^{s}$ is extracted as the scene representation through the same layers' operation as nodes. {For directed edge $e_{j\to i}$ from node $v_{j}$ to $v_{i}$, we use both the spatial feature and visual feature of $v_i, v_j$ to compute a scalar, which represents the influence of $v_j$ on $v_i$, as will be detailed in Sec. \ref{sec:SI}}. { With such modeling, how to drive them to interact in the graph? It will be delineated in the following.} 

\subsection{Message Passing}

\begin{figure}[t]
\vspace{-2ex}
\begin{center}
   \includegraphics[width=0.97\linewidth]{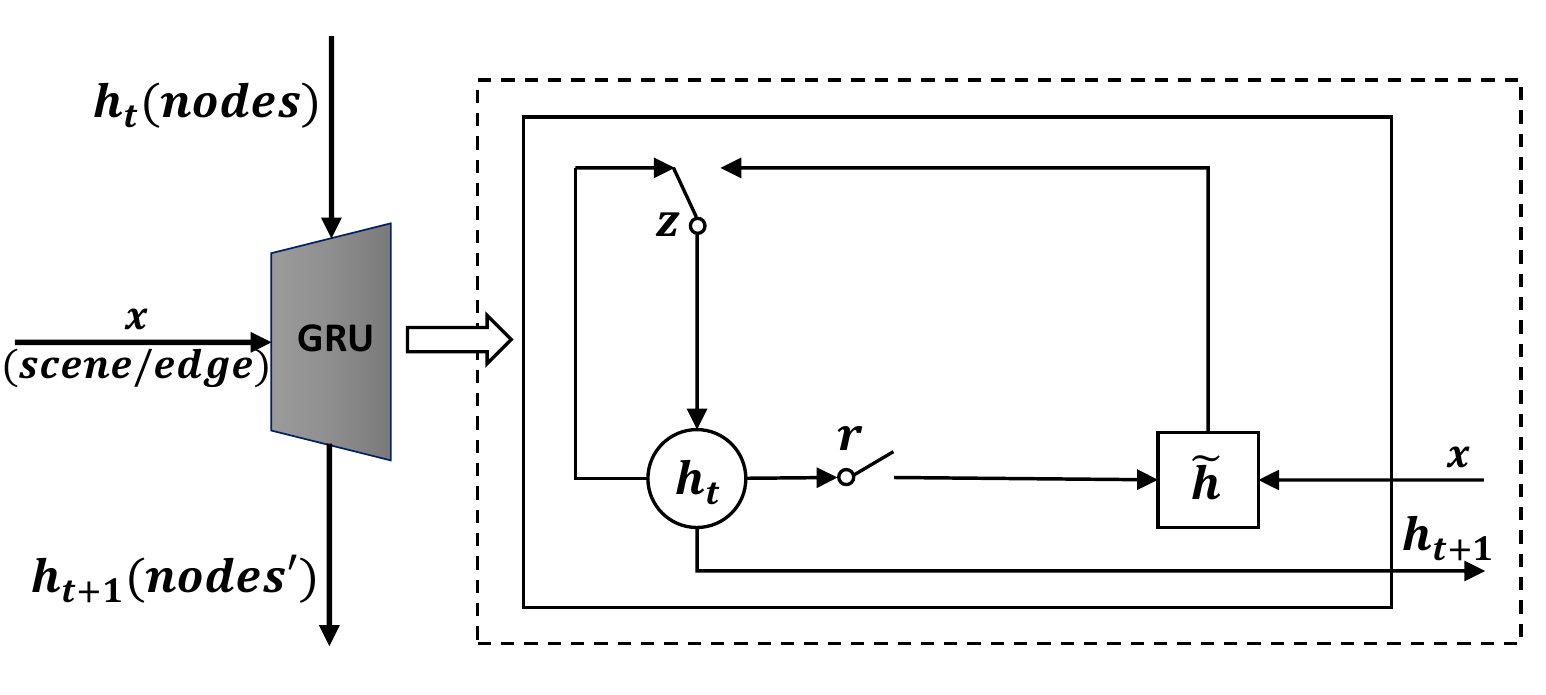}
\end{center}
\vspace{-2ex}
   \caption{{\bf An illustration of GRU.} The update gate $z$ selects whether the hidden state $h_{t+1}$ is to be updated with a new hidden state $\tilde{h}$. The reset gate $r$ decides whether the previous hidden state $h_t$ is ignored.} 
\label{fig:GRU}
\label{fig:onecol}
\vspace{-2.5ex}
\end{figure}

For each node, the key of interaction is to encode the messages passed from the scene and other nodes to it. Due to that each node needs receiving multiple incoming messages, it is necessary to design an aggregation function that can remember the node details itself and then fuse incoming messages into a meaningful representation. Considering this function behaves like a memory machine, we explore RNNs. As is well known, an RNN can in principle map from the entire history of previous inputs to each outputs. The key point is that the recurrent connections allow a memory of previous inputs to persist in the network's internal state, and thereby influence the network output \cite{Book}. Since that GRU\cite{GRU} as a special kind of RNN is lightweight and effective, it is used to act like memory machines in this work.

Let us review how a GRU cell works in Fig. \ref{fig:GRU}. First, the \emph{reset} gate $r$ is computed by
\begin{equation}
r = \sigma(W_r[x, h_{t}]) ,
\end{equation}
where $\sigma$ is the logistic sigmoid function, and [,] denotes the concatenation of vectors. $W_r$ is a weight matrix which is learned. $h_t$ is the previous hidden state, by the way, the input $x$ and $h_t$ have the same dimensions. 
Similarly, the \emph{update} gate $z$ is computed by
\begin{equation}
z = \sigma(W_z[x, h_{t}]) .
\end{equation}
The actual activation of the proposed unit $h_{t+1}$ is then computed by
\begin{equation}
h_{t+1} = zh_t + (1-z)\tilde{h} ,
\end{equation}
where
\begin{equation}
\tilde{h}=\phi(Wx + U(r\odot h_t)) .
\end{equation}
$\phi$ denotes $tanh$ activate function, $W$ and $U$ are weight matries which are learned. $\odot$ denotes the element-wise multiplication. As stated in \cite{GRU}, in the above formulations, the memory cell allows the hidden state to drop any information that is found to be irrelevant with input later through the reset gate $r$. On the other hand, the memory cell can control how much information from the previous state will carry over to the current hidden state, thus, allowing a more compact representation through the update gate $z$. 

{Generally, GRU as an effective memory cell can remember long-term information, where the initial state of GRU is empty or a random vector and the input is a sequence of symbols. In this paper, we use GRU to encode different kinds of messages to object state. {\bf To encode message from scene}, we take the fine-grained object details as initial state of GRU, and take the message from scene as input. GRU cell could choose to ignore some parts of object state which are not relative with this scene context, or use scene context to enhance some parts of object state. {\bf To encode message from other objects}, we also take the object details as initial state of GRU, and take an integrated message from other nodes as input. The memory cell would also play a same role to choose relative information to update the hidden state of objects. When the state of object updated, the relationships among objects will also change, then more time steps of updating make the model more stable.}

\subsection{Structure Inference} \label{sec:SI}
\begin{figure}[t]
\begin{center}
   \includegraphics[width=1.0\linewidth, trim=10 20 10 20, clip]{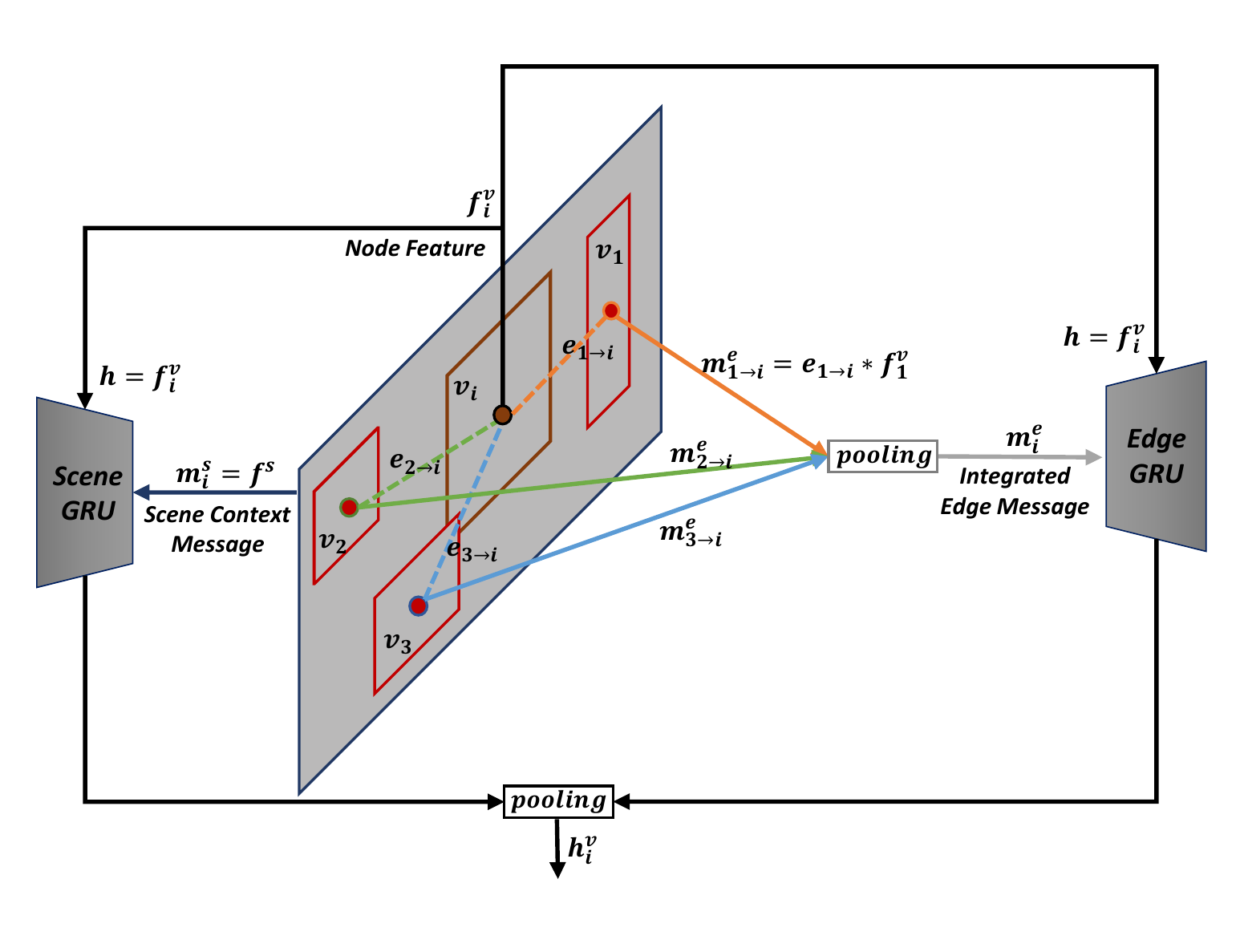}
\end{center}
\vspace{-2ex}
   \caption{{\bf Structure Inference.} For object $v_i$, the input of scene GRU is scene context message $m^s_i$, and the initial hidden state is the node $v_i$ feature $f^v_i$. For message $m^e_{1\to i}$ from node $v_1$ to node $v_i$ is controlled by edge $e_{1\to i}$. These messages from all other objects are integrated as $m^e_i$ to input the edge GRU. The initial hidden state of edge GRU is also $f^v_i$. Then these two sets of GRU output ensemble together as eventual updated node state.}
\label{fig:SI}
\vspace{-2ex}
\end{figure}


\begin{table*}
\caption{{\bf Detection Results on VOC 2007 test.} Legend: {\bf 07+12:} 07 trainval + 12 trainval.}
\vspace{1ex}
\label{table:voc}
\begin{center}
\resizebox{\textwidth}{!} {
\begin{tabular}{l|c|c|cccccccccccccccccccc}
\hline
Method & Train & $mAP$ & aero & bike & bird  & boat & bottle & bus & car & cat & chair & cow & table & dog & horse & mbike & person & plant & sheep & sofa & train & tv \\
\hline
\hline
Fast R-CNN \cite{Fast} & 07+12 & 70.0 & 77.0 & 78.1 & 69.3 & 59.4 & 38.3 & 81.6 & 78.6 & 86.7 & 42.8 & 78.8 & 68.9 & 84.7 & 82.0 & 76.6 & 69.9 & 31.8 & 70.1 & {\bf 74.8} & 80.4 & 70.4 \\
Faster R-CNN \cite{Faster} & 07+12 & 73.2 & 76.5 & 79.0 & 70.9 & 65.5 & 52.1 & 83.1 & 84.7 & 86.4 & 52.0 & 81.9 & 65.7 & 84.8 & 84.6 & 77.5 & 76.7 & 38.8 & 73.6 & 73.9 & 83.0 & 72.6 \\ 
SSD500 \cite{SSD} & 07+12 & 75.1 & {\bf 79.8} & 79.5 & 74.5 & 63.4 & 51.9 & 84.9 & 85.6 & 87.2 & 56.6 & 80.1 & 70.0 & 85.4 & 84.9 & 80.9 & 78.2 & {\bf 49.0} & {\bf 78.4} & 72.4 & {\bf 84.6} & 75.5 \\
ION \cite{ION} & 07+12 & 75.6 & 79.2 & {\bf 83.1} & {\bf 77.6} & 65.6 & 54.9 & {\bf 85.4} & 85.1 & 87.0 & 54.4 & 80.6 & {\bf 73.8} & 85.3 & 82.2 & {\bf 82.2} & 74.4 & 47.1 & 75.8 & 72.7 & 84.2 & {\bf 80.4} \\
SIN (ours) & 07+12 & {\bf 76.0} & 77.5 & 80.1 & 75.0 & {\bf 67.1} & {\bf 62.2} & 83.2 & {\bf 86.9} & {\bf 88.6} & {\bf 57.7} & {\bf 84.5} & 70.5 & {\bf 86.6} & {\bf 85.6} & 77.7 & {\bf 78.3} & 46.6 & 77.6 & 74.7 & 82.3 & 77.1 \\
\hline
\end{tabular}}
\end{center}
\vspace{-3ex}
\end{table*}

\begin{table*}
\caption{{\bf Detection Results on VOC 2012 test.} Legend: {\bf 07++12:} 07 trainval + 12 trainval + 07 test.}
\vspace{-1ex}
\label{table:voc12}
\begin{center}
\resizebox{\textwidth}{!} {
\begin{tabular}{l|c|c|cccccccccccccccccccc}
\hline
Method & Train & $mAP$ & aero & bike & bird & boat & bottle & bus & car & cat & chair & cow & table & dog & horse & mbike & person & plant & sheep & sofa & train & tv \\
\hline
\hline
Fast R-CNN \cite{Fast} & 07++12 & 68.4 & 82.3 & 78.4 & 70.8 & 52.3 & 38.7 & 77.8 & 71.6 & {89.3} & 44.2 & 73.0 & 55.0 & {87.5} & 80.5 & 80.8 & 72.0 & 35.1 & 68.3 & 65.7 & 80.4 &64.2 \\
SSD300 \cite{SSD} & 07++12 & 70.3 & 84.2 & 76.3 & 69.6 & 53.2 & 40.8 & {78.5} & 73.6 & 88.0 & 50.5 & 73.5& {\bf 61.7} & 85.8 & 80.6 & 81.2 & 77.5 & 44.3 & 73.2 & {\bf 66.7} & 81.1 & 65.8 \\
Faster R-CNN \cite{Faster} & 07++12 & 70.4 & {\bf 84.9} & {\bf 79.8} & 74.3 & 53.9 & 49.8 & 77.5 & 75.9 & 88.5 & 45.6 & {\bf 77.1} & 55.3 & 86.9 & 81.7 & 80.9 & 79.6 & 40.1 & 72.6 & 60.9 & {81.2} & 61.5 \\
HyperNet \cite{Hyper} & 07++12 & 71.4 & 84.2 & 78.5 & 73.6 & 55.6 & 53.7 & 78.7 & {\bf 79.8} & 87.7 & 49.6 & 74.9 & 52.1 & 86.0 & 81.7 & {\bf 83.3} & {\bf 81.8} & {48.6} & 73.5 & 59.4 & 79.9 & 65.7 \\
SIN (ours) & 07++12 & {\bf 73.1} & {84.8} & 79.5 & {\bf 74.5} & {\bf 59.7} & {\bf 55.7} & {\bf 79.5} & 78.8 & {\bf 89.9} & {\bf 51.9} & {76.8} & 58.2 & {\bf 87.8} & {\bf 82.9} & 81.8 & 81.6 & {\bf 51.2} & {\bf 75.2} & 63.9 & {\bf 81.8} & {\bf 67.8} \\

\hline
\end{tabular}}
\end{center}
\vspace{-3ex}
\end{table*}

\begin{table*}[!htb]
\caption{{\bf Detection Results on COCO 2015 test-dev.} Legend: {\bf trainval35k:} COCO train + 35k val. {\bf *}Baseline our trained.}
\vspace{-1ex}
\begin{center}
\resizebox{\textwidth}{!}{
\begin{tabular}{l|c|ccc|ccc|ccc|ccc}
\hline
Method &  Train & $AP$ & $AP^{50}$ & $AP^{70}$ & $AP^{S}$ & $AP^{M}$ & $AP^{L}$ & $AR^{1}$ & $AR^{10}$ & $AR^{100}$ & $AR^{S}$ & $AR^{M}$ & $AR^{L}$\\
\hline
\hline
Fast R-CNN \cite{Fast} & train & 20.5 & 39.9 & 19.4 & 4.1 & 20.0 & {35.8} & 21.3 & 29.5 & 30.1 & 7.3 & 32.1 & 52.0 \\
Faster R-CNN* \cite{Faster} & train & 21.1 & 40.9 & 19.9 & 6.7 & 22.5 & 32.3 & 21.5 & 30.4 & 30.8 & 9.9 & 33.4 & 49.4\\

YOLOv2 \cite{YOLO9000} & trainval35k \cite{ION} & 21.6 & 44.0 & 19.2 & 5.0 & 22.4 & 35.5 & 20.7 & {31.6} & {\bf 33.3} & 9.8 & {36.5} & {\bf 54.4} \\
ION \cite{ION} & train & 23.0 & 42.0 & {\bf 23.0} & 6.0 & 23.8 & {\bf 37.3} & {\bf 23.0} & {\bf 32.4} & 33.0 & 9.7 & {\bf 37.0}  & 53.5\\
SIN (ours) & train & {\bf 23.2} & {\bf 44.5} & {22.0} & {\bf 7.3} & {\bf 24.5} & {36.3} & {22.6} & {31.6} & {32.0} & {\bf 10.5} & {34.7} & {51.3} \\
\hline
\end{tabular}}
\end{center}
\label{table:coco}
\vspace{-4ex}
\end{table*}

To encode two kinds of messages above, a set of scene GRUs and edge GRUs are designed to propagate message from scene and other objects to node. Then nodes are updated according to the graph, as shown in Fig. \ref{fig:SI}.

The scene GRU takes nodes visual feature $f^{v}$ as initial hidden states, and takes scene message $m^{s}$ as input, which is exactly scene context $f^s$ as shown in the left part of Fig. \ref{fig:SI}. As described above, the scene GRU would learn its key gates function to choose information to update nodes.

The edge GRU is used to encode messages from many other objects, there we need to calculate an integrated message $m_e$ in advance, or we need take a long sequence of messages from every other object as inputs, which will cost very much. For each node, the edge GRU will choose parts of the integrate message to update this node. For the messages passed from other objects to node $v_i$, various objects contribute differently. So we model every object-object relationship $e_{j\to i}$ as a scalar weight, which represents the influence of $v_j$ on $v_i$. It is reasonable that object-object relationship $e_{j\to i}$ is common determined by relative object position and visual clues, {\em e.g.} a mouse is more important to the keyboard than a cup and more close mouse is more important to the keyboard. As shown in the right part of Fig. \ref{fig:SI}, the integrated message to node $v_i$ is calculated by 
\begin{equation}
m^e_{i} = \max_{j \in V} pooling(e_{j \to i}* f^v_j) ,
\vspace{-1ex}
\end{equation}
where
\vspace{-1ex}
\begin{equation}
e_{j\to i} = relu(W_pR^p_{j\to i}) * tanh(W_v[f^v_i, f^v_j]) .
\end{equation}
$W_p$ and $W_v$ are learnable weight matrixes. Using max-pooling can extract the most important message, while if using mean-pooling, message might be disturbed by the large number of ROIs from irrelevant regions. The visual relationship vector is formed by concatenating visual feature $f^v_i$ and $f^v_j$. $R^p_{j\to i}$ denotes the spatial position relationship, which is represented as 
\begin{equation}
\begin{split}
R^p_{j \to i} = & [w_i, h_i, s_i, w_j, h_j, s_j, \frac{(x_i - x_j)}{w_j} ,\frac{(y_i - y_j)}{h_j}, \\& \frac{(x_i - x_j)^2}{w_j^2}, \frac{(y_i - y_j)^2}{h_j^2}, log(\frac{w_i}{w_j}), log(\frac{h_i}{h_j})],
\end{split}
\end{equation}
where $(x_i, y_i)$ is the center of ROI $b_i$, while $w_i, h_i$ are the width and height of $b_i$, and $s_i$ is the area of $b_i$.

For node $v_{i}$, it receives messages both from the other nodes and scene context. Eventually we get the comprehensive representation $h_{t+1}$, which denotes the node state. In our current study, we empirical find that (details in Sec. \ref{sec:way}) mean-pooling is the most effective, compared to max-pooling and concatenation, so
\begin{equation}
h_{t+1} =\frac{h^s_{t+1}+ h^e_{t+1}}{2},
\end{equation}
where $h^s_{t+1}$ is the output of scene GRU, and $h^e_{t+1}$ denotes the output of edge GRU.

In the following iterations, scene GRUs will put the new (updated) node state as their hiddens, and take fixed scene feature as input, then compute next node states. Edge GRUs would take the new object-object message as new input, then compute the next hidden states. Finally, the eventual integrated node representations are used to predict object category and bounding box offsets.


\section{Results}

In this part, we comprehensively evaluate SIN on two datasets including PASCAL VOC \cite{voc} and MS COCO \cite{coco}. 


\subsection{Implementation Details}
{We use a VGG-16 model pre-trained on ImageNet \cite{imagenet}. During training and testing stage, we use NMS \cite{NMS} to select 128 boxes as object proposals. Faster R-CNN is trained by ourself as {\em baseline}, where all parameters are set according to the original publications. For our method, since we find that smaller learning rate is more suitable, consequently the number of train iterations is increased. The momentum, weight decay and batch size are all the same as {\em baseline}. Specifically, when training on VOC 2007 trainval combined with VOC 2012 trainval and testing on VOC 2007 test, we use a learning rate of $5\times10^{-4}$ for 80k iterations, and $5\times10^{-5}$ for the next 50k iterations. When training on VOC 2007 trainvaltest combined with VOC 2012 trainval and testing on VOC 2012 test, we use a learning rate of $5\times10^{-4}$ for 100k iterations, and $5\times10^{-5}$ for the next 70k iterations. When training on COCO train and testing on COCO 2015 dev-test, we use a learning rate of $5\times10^{-4}$ for 350k mini-batches, and $5\times10^{-5}$ for the next 200k mini-batches. Our method and {\em baseline} are both implemented with Tensorflow\footnote {Our source code is available at \href{http://vipl.ict.ac.cn/resources/codes}{http://vipl.ict.ac.cn/resources/codes}.} {\cite{TF}}.

\subsection{Overall Performance}
{\bf PASCAL VOC.} VOC involves 20 categories. VOC 2007 dataset consists of about 5k trainval images and 5k test images, while VOC 2012 dataset includes about 11k trainval images and 11k test images. We set two kinds of train dataset, and the evaluations were carried out on the VOC 2007 and VOC 2012 test set (from VOC 2012 evaluation server) respectively in Tab. \ref{table:voc} and Tab. \ref{table:voc12}. Applying our method, we get a higher mAP of 76.0\% on VOC 2007 and a mAP of 73.1\% on VOC 2012 test. Especially to deserve to be mentioned, our method is also better than ION \cite{ION} on VOC 2007 test, which is a multi-scale network with explicit modeling of context using a recurrent network. 

{\bf MS COCO.} To further validate our method on a larger and more challenging dataset, we conduct experiments on COCO and report results from test-dev 2015 evaluation server in Tab. \ref{table:coco}. The evaluation metric of COCO dataset is different from VOC. The overall performance $AP$ averages $mAP$ over different IOU thresholds from 0.5 to 0.95. This places a significantly larger emphasis on localization compared to the VOC metric with only requires IOU of 0.5. In this more challenging dataset, our SIN achieves 23.2\% on test-dev score, again verifying the advantage of our method.

\section{Design Evaluation}
\label{sec:study}
In this section, we explore the effectiveness of our model, including two main modules of using scene contextual information named as {\em Scene} and using object relative relationships named as {\em Edge}. Additionally, we conduct in-depth analysis of the performance metrics of our method.

\begin{table*}
\caption{{\bf Ablation Study on VOC 2007 test.} All methods are trained on VOC 2007 trainval. {\bf Baseline}: Faster R-CNN our trained. {\bf {\em Scene}}: only using scene context. {\bf {\em Edge:}} only using object-object relationships.}
\vspace{-1ex}
\label{table:vocStudy}
\begin{center}
\resizebox{\textwidth}{!} {
\begin{tabular}{l|c|cccccccccccccccccccc}
\hline
Method & $mAP$ & aero & bike & bird & boat & bottle & bus & car & cat & chair & cow & table & dog & horse & mbike & person & plant & sheep & sofa & train & tv \\
\hline
\hline
Baseline & 68.79 & 68.86 & 77.70 & 67.52 & 54.00 & 53.84 & 75.98 & \textbf{80.07} & 79.89 & 49.31 & 73.98 & 65.80 & 77.15 & 80.21 & 76.52 & 76.88 & 38.72 & 66.75 & 65.48 & 75.54 & 71.53\\
{\em Scene} & 70.23 & \textbf{70.11} & \textbf{78.38} & \textbf{69.33} & \textbf{60.88} & 53.09 & 76.98 & 79.64 & \textbf{86.01} & 49.86 & \textbf{75.02} & 68.00 & \textbf{78.66} & \textbf{80.66} & 74.70 & 77.34 & \textbf{41.21} & 68.28 & 65.38 & \textbf{76.59} & 74.47\\
{\em Edge} & \textbf{70.31} & 70.08 & 78.20 & 67.46 & 57.64 & \textbf{56.04} & \textbf{78.54} & 80.02 & 79.89 & \textbf{51.10} & 74.12 & \textbf{70.17} & 77.99 & 80.58 & \textbf{77.54} & \textbf{77.60} & 41.07 & \textbf{69.04} & \textbf{68.33} & 76.20 & \textbf{74.60} \\
\hline
\end{tabular}}
\end{center}
\vspace{-4.5ex}
\end{table*}

\begin{table}
\caption{{\bf Ablation Study on COCO test-dev 2015.} All methods are trained on COCO train set. {\bf Baseline}: Faster R-CNN our trained. {\bf {\em Scene}}: only using scene context. {\bf {\em Edge}}: only using object-object relationships.}
\begin{center}
\vspace{-2.5ex}
\resizebox{0.48\textwidth}{!}{
\begin{tabular}{l|ccc|ccc}
\hline
Method & $AP$ & $AP^{50}$ & $AP^{70}$ & $AP^{S}$ & $AP^{M}$ & $AP^{L}$ \\
\hline
\hline
Baseline & 21.1 & 40.9 & 19.9 & 6.7 & 22.5 & 32.3\\
\em{Scene}  & 22.5 & {\bf 43.9} & 21.1 & {\bf 7.1} & 24.1 & 34.9 \\
\em{Edge}  & {\bf 22.7}  & 43.3 & {\bf 21.6} & 7.0  & {\bf 24.2} & {\bf 35.7}\\
\hline
\end{tabular}}
\end{center}
\label{table:cocoRes}
\vspace{-2.5ex}
\end{table}

\subsection{Scene Module}
In this experiment, only scene contextual information is considered to update nodes feature. In other words, just a set of scene GRUs is used in structure inference. 

{\bf Performance.} As shown in Tab. \ref{table:vocStudy}, for the simplify to do ablation study, all methods are trained on VOC 2007 trainval and test on VOC 2007 test. {\em Scene} module achieves a better mAP of 70.23\% compared with baseline on VOC. Interestingly, it is found that {\em Scene} gets a prominent average precision on some categories including {\em aeroplane}, {\em bird}, {\em boat}, {\em table}, {\em train}, {\em tv} and so on, especially the average precision of {\em boat} increases by more than 6\%. This result is actually not surprising since one can find that such categories generally have pretty high correlations with the scene context. For instance, planes and birds are mostly in the sky, while boats are commonly in the river.

\begin{figure}[t]
\centering
 \includegraphics[width=1.0\linewidth,  trim=240 190 240 190, clip
 ]{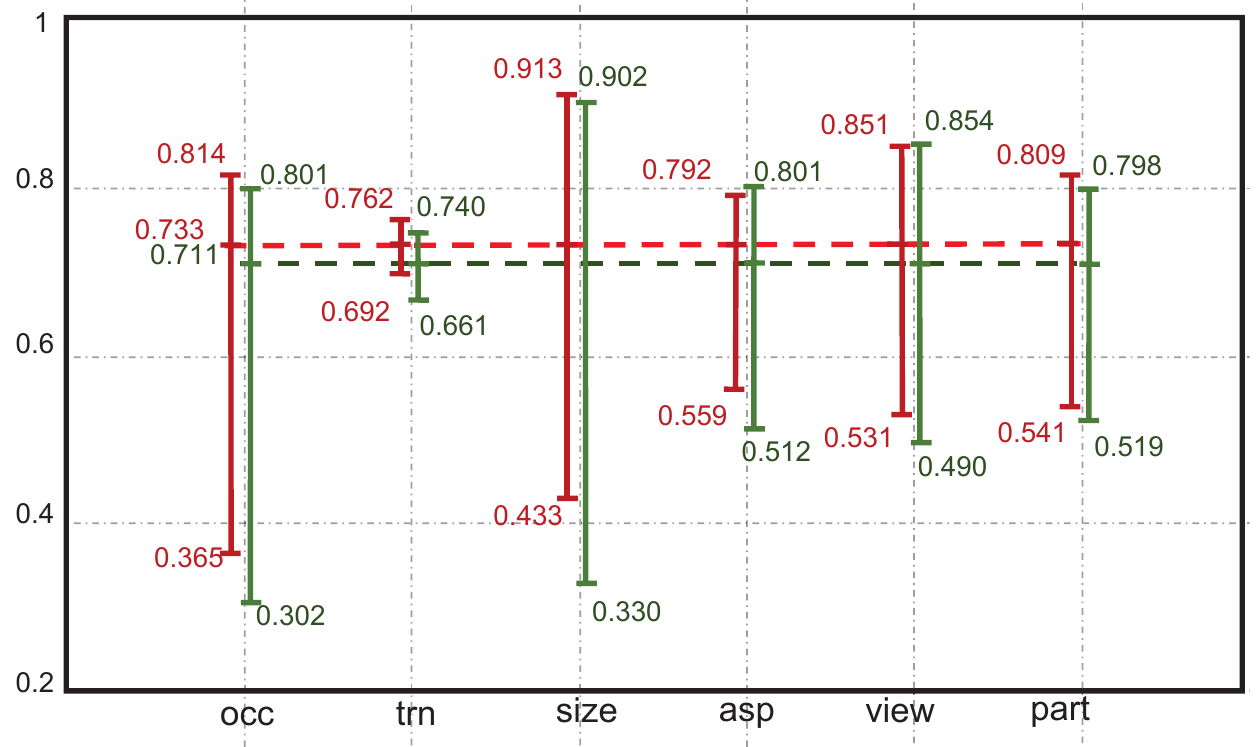}
    \caption{{\bf Summary of Sensitivity and Impact of Object Characteristics.} We show the average (over 7 categories) Normalized AP($AP_N$ \cite{Error}) of the highest performing and lowest performing subsets within each characteristic (occlusion, truncation, bounding box area, aspect ratio, viewpoint, part visibility). Overall $AP_N$ is indicated by the dashed line. The difference between max and min indicates sensitivity. The difference between max and overall indicates the impact. {\bf Red}: {\em Scene}. {\bf Green}: baseline.}
\label{fig:impact}

\end{figure}

\begin{figure}[t]
\centering
\vspace{-2ex}
\hspace{-3ex}
\subfigure {
\label{O}
\includegraphics[width=1.0\linewidth,  trim=50 600 50 85, clip
]{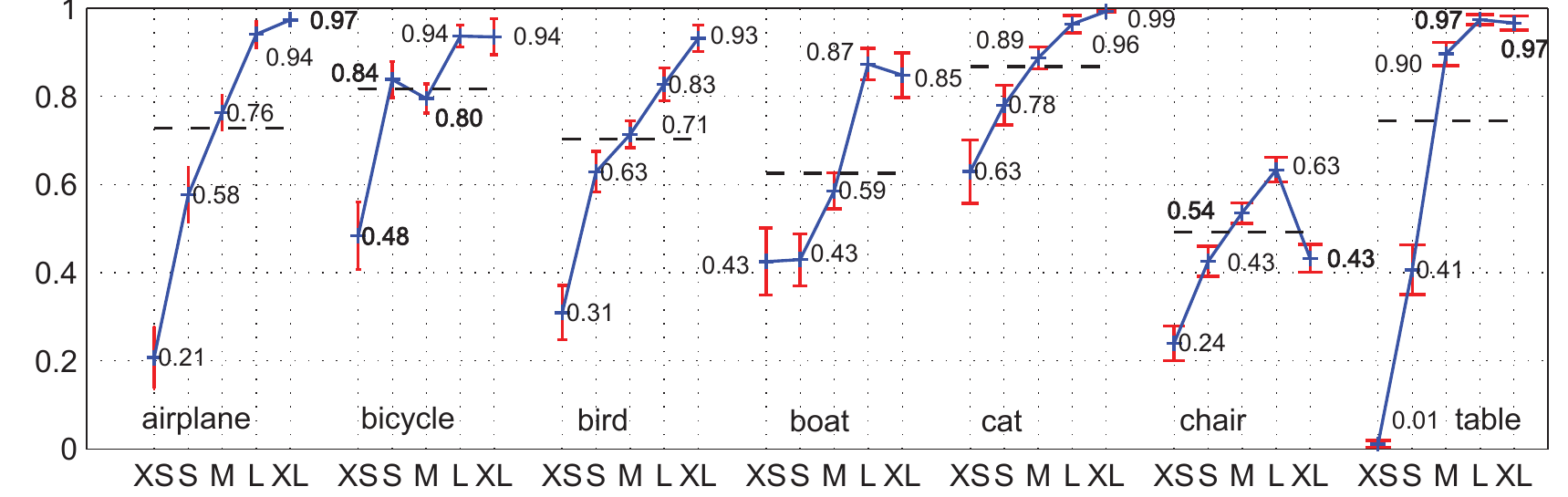}}\\
\hspace{-3ex}
\subfigure {
\label{O}
 \includegraphics[width=1.0\linewidth, trim=50 600 50 85, clip]{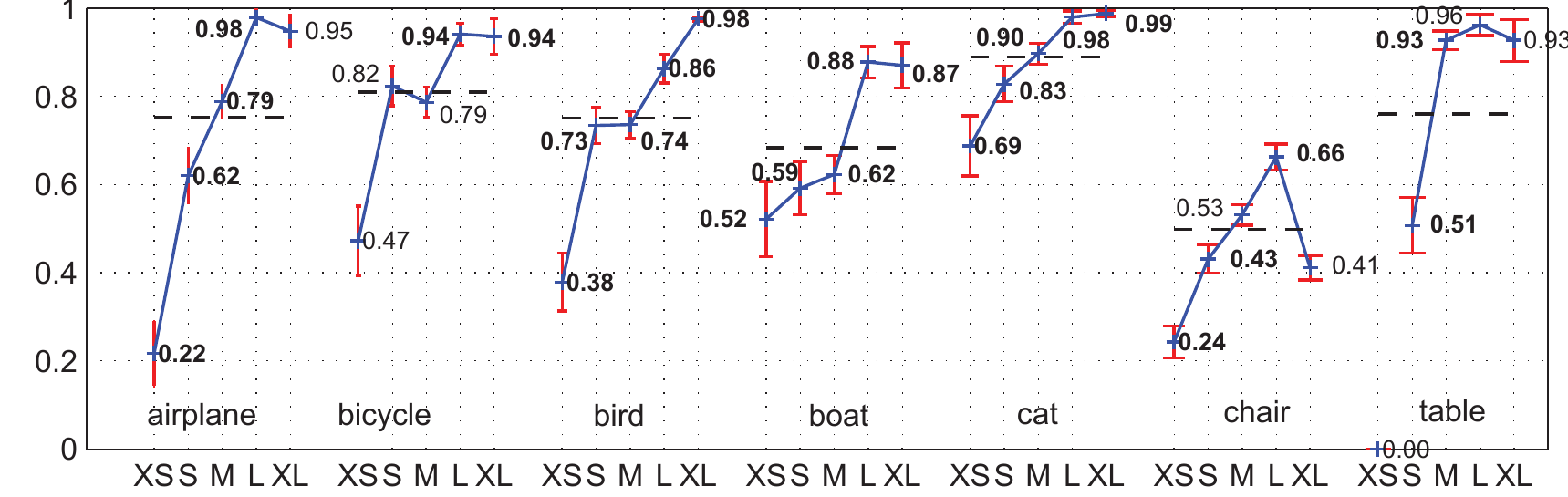}}
\caption{{\bf Sensitivity and Impact of BBox Area on VOC 2007 test.} Each plot shows $AP_N$ \cite{Error} with standard error bars (red). Black dashed lines indicate overall $AP_N$. The plot shows the effects of BBox Area per category. Key: BBox Area: {\bf XS}=extra-small; {\bf S}=small; {\bf M}=medium; {\bf L}=large; {\bf XL}=extra-large. The {\bf top} figure is for {\em baseline}, and the {\bf bottom} one is for {\em Scene}. }
\label{fig:area}
\vspace{-2ex}
\end{figure}

{\bf Small, vague or occluded object.} To further examine the differences between baseline and {\em Scene}, we look at a detailed breakdown of results of VOC 2007. We use the detection analysis tool from \cite{Error}. Fig. \ref{fig:impact} provides a compact summary of the sensitivity to each characteristic and the potential impact of improving robustness on seven categories selected by \cite{Error}. Overall, our method is more robust than baseline against occlusion, truncation, area size and part. Efforts to improve these characteristics are explicit. The further specialized analysis on area size is shown in Fig. \ref{fig:area}. Our method gets a distinct improvement on extra-small bird, boat and cat category, and achieves better performance on other size. Besides, the $AP^S$ of COCO test depicted in Tab. \ref{table:cocoRes} which represents the performance of small objects also gets improved compared with baseline. 

{\bf Qualitative results of \em Scene.} Additionally, a couple of examples of how {\em Scene} module can help improve the detection performance are shown in Fig. {\ref{fig:sceneRes}}. In the first case, some {\em boats} are mislabeled as {\em car} by the baseline of Faster R-CNN, while our method correctly labeled these vague objects as {\em boats}. In the second case, nothing is detected by the baseline, however a {\em chair} is detected using scene contextual information. The third one is a failure case, where an {\em aeroplane} is truly detected in a quite rare situation (on the river) by the baseline but it is misclassified as a {\em boat} by our model. This sample suggests us further improve our method to flexibly balance the general cases and rare ones by weighting the importance of global scene context.

\begin{figure}[t]
\vspace{-2ex}
\centering
\subfigure[mislabeled {\em boats} ] {
 \label{O}
 \includegraphics[width=0.31\linewidth]{001188_o.jpg}} 
 \hspace{-1ex}
\subfigure[nothing detected] {
 \label{O}
 \includegraphics[width=0.3\linewidth]{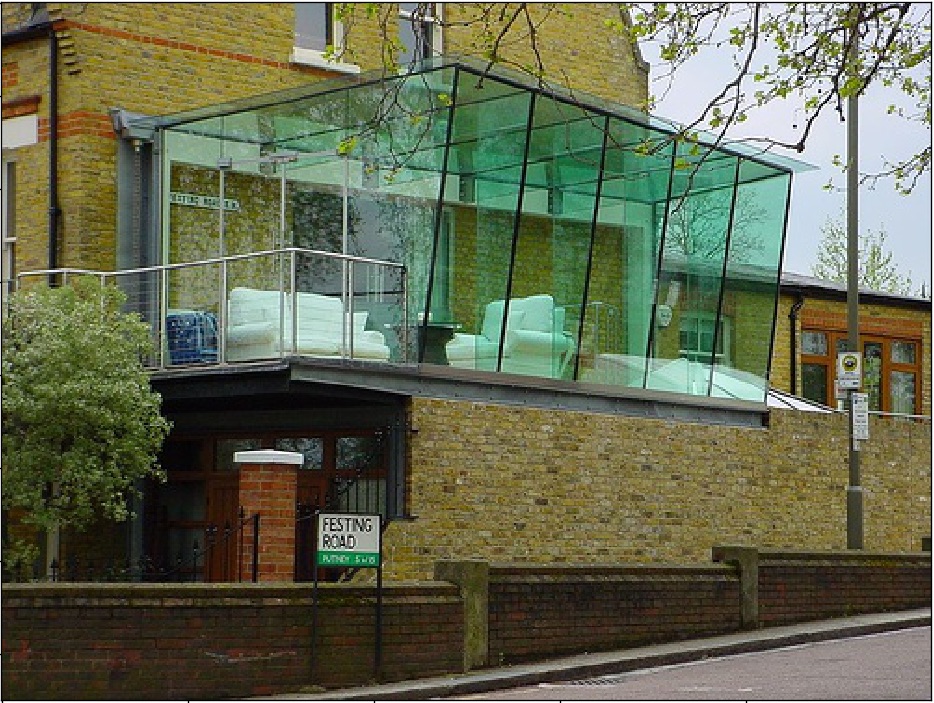}}
 \subfigure[an {\em aero} with a {\em boat}] {
 \label{O}
 \includegraphics[width=0.34\linewidth]{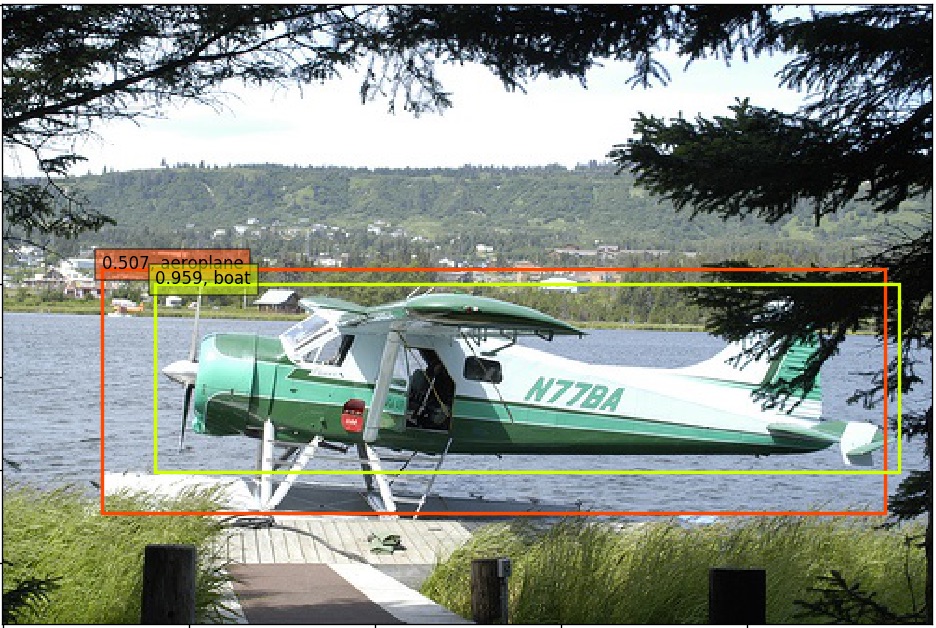}} \\
 \vspace{-2ex}
\subfigure[{\em boats} are detected] {
 \label{O}
 \includegraphics[width=0.31\linewidth]{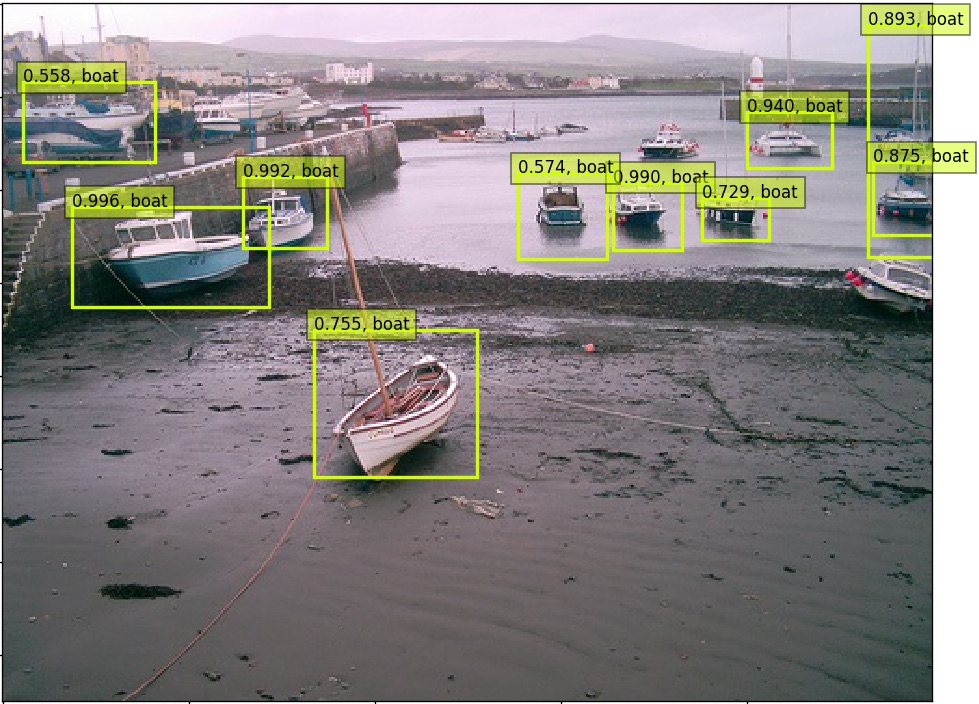}}
 \hspace{-1ex}
  \subfigure[{\em chair} is detected] {
 \label{O}
 \includegraphics[width=0.3\linewidth]{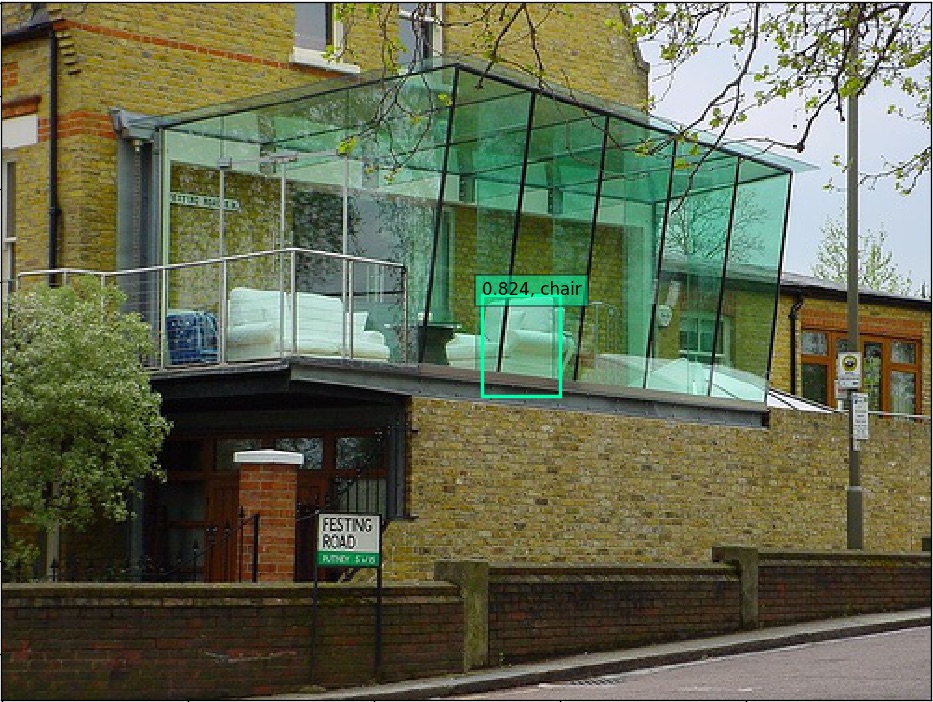}}
  \subfigure[only a {\em boat}] {
 \label{O}
 \includegraphics[width=0.34\linewidth]{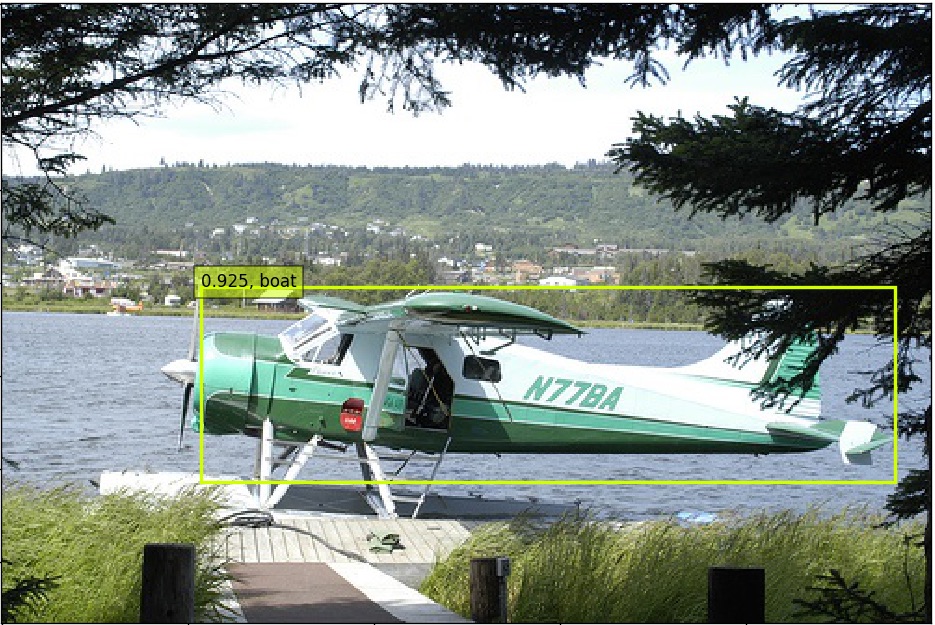}} \\

\caption{{\bf Qualitative results of Baseline vs. {\em Scene} on VOC.} In every pair of detection results (top vs. bottom), the top is based on baseline, and the bottom is detection result of {\em Scene.}}
\label{fig:sceneRes}
\vspace{-2ex}
\end{figure}

\subsection{Edge Module}
We evaluate the effectiveness of only {\em Edge} module in this part. Like {\em Scene} module, only a set of edge GRUs is used to direct the nodes updating according to relative objects. From Tab. \ref{table:vocStudy} and \ref{table:cocoRes}, its advantage over the baseline is again verified.

{\bf Localization.} To understand the effectiveness of {\em Edge} in more details, we use the detection analysis tool in \cite{Error} again. 
It is found that most categories have enjoyed more accurate localization compared with the baseline. Fig. \ref{fig:loc} takes two example categories (i.e., {\em aeroplane} and {\em bus}) to show the frequency and impact on the performance of each type of false positive. One can see that the localization error has been largely decreased. More results are provided in supplementary material. By further checking the results of COCO in Tab. \ref{table:cocoRes}, the $AP^{70}$ improves greatly, which means that our method {{provides more accurate results.}}

\begin{figure}[t]
\vspace{-1ex}
\centering
\subfigure[{\em aeroplane} on baseline] {
 \label{O}
 \includegraphics[width=0.38\linewidth,  trim=40 635 410 65, clip]{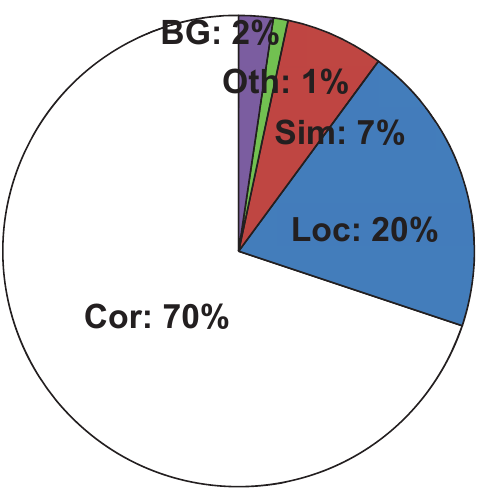}} 
\subfigure[{\em aeroplane} on {\em Edge}] {
 \label{O}
 \includegraphics[width=0.38\linewidth, trim=40 635 410 65, clip]{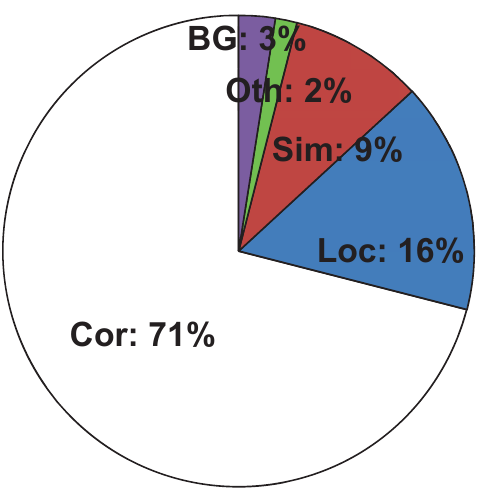}}\\
 \vspace{-2ex}
\subfigure[{\em bus} on baseline ] {
 \label{O}
 \includegraphics[width=0.38\linewidth, trim=40 635 410 65, clip]{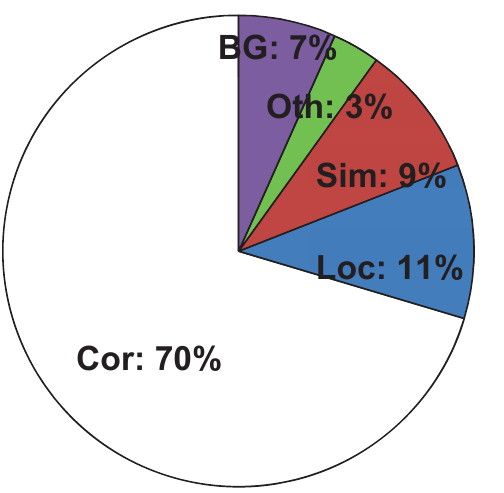}}
  \subfigure[{\em bus} on {\em Edge}] {
 \label{O}
 \includegraphics[width=0.38\linewidth, trim=40 635 410 65, clip]{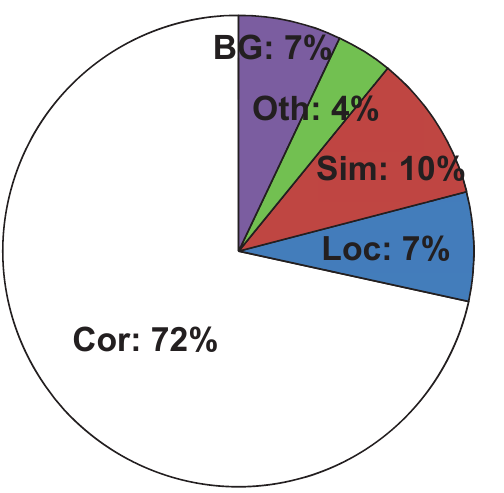}} \\ 
 
\caption{{\bf Analysis of Top-Ranked False Positives.} Pie charts: fraction of detections that are correct (Cor) or false positive due to poor localization (Loc), confusion with similar objects (Sim), confusion with other VOC objects (Oth), or confusion with background or unlabeled objects (BG). {\bf Left}: results of the baseline Faster R-CNN. {\bf Right}: results of {\em Edge}. Loc errors are fewer than {\em baseline} on {\em aeroplane} and {\em bus}.}
\label{fig:loc}
\vspace{-2.5ex}
\end{figure}

{\bf Qualitative results of {\em Edge}.} Comparing qualitative results between baseline and {\em Edge} module, we find a common type of detection error of Faster R-CNN that one object would be detected by two or more boxes labeled as similar categories, because Faster R-CNN predicts a specific regression box for each possible category given a candidate region. It would record all high score categories with the specific boxes. Namely, one candidate box would produce numbers of close detection results. As shown in Fig. \ref{fig:edgeRes}(a)(c), the multiple box results of one object detected by baseline are redundant. This kind of errors can be largely reduced by {\em Edge}, due to that object relationships between those overlapping nodes make them homogenized. Not only a higher accuracy is achieved, detection results also look more comfortable by using {\em Edge} in Fig. \ref{fig:edgeRes}(b)(d).

\begin{figure}[t]
\vspace{-1ex}
\centering
\subfigure[{\em cars} are redundant] {
 \label{O}
 \includegraphics[width=0.47\linewidth]{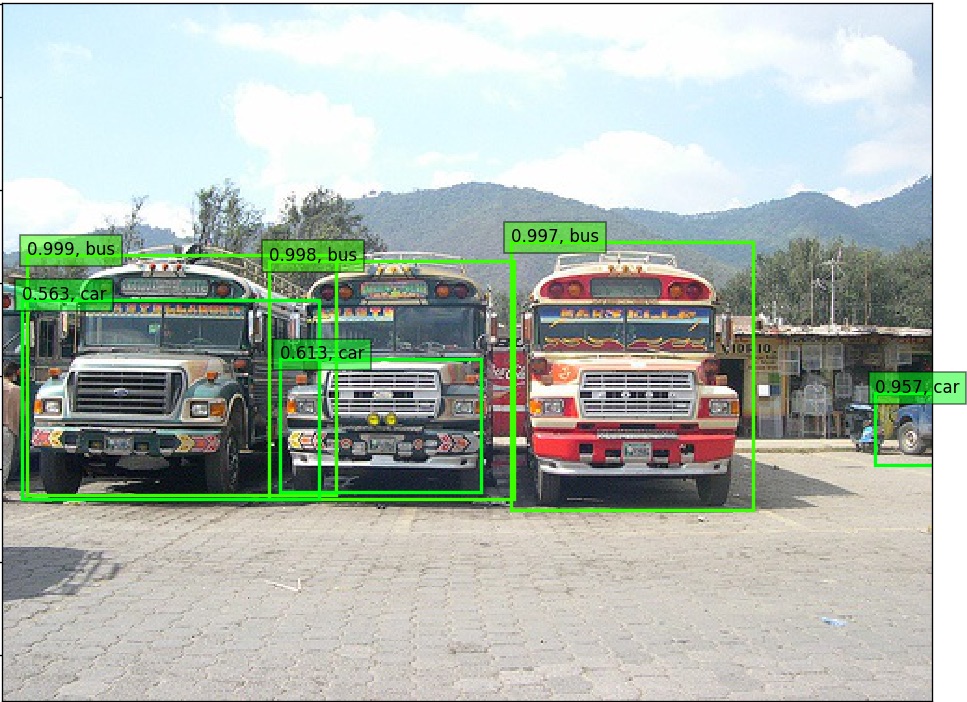}} 
 \hspace{-1ex}
\subfigure[results of {\em Edge}] {
 \label{O}
 \includegraphics[width=0.47\linewidth]{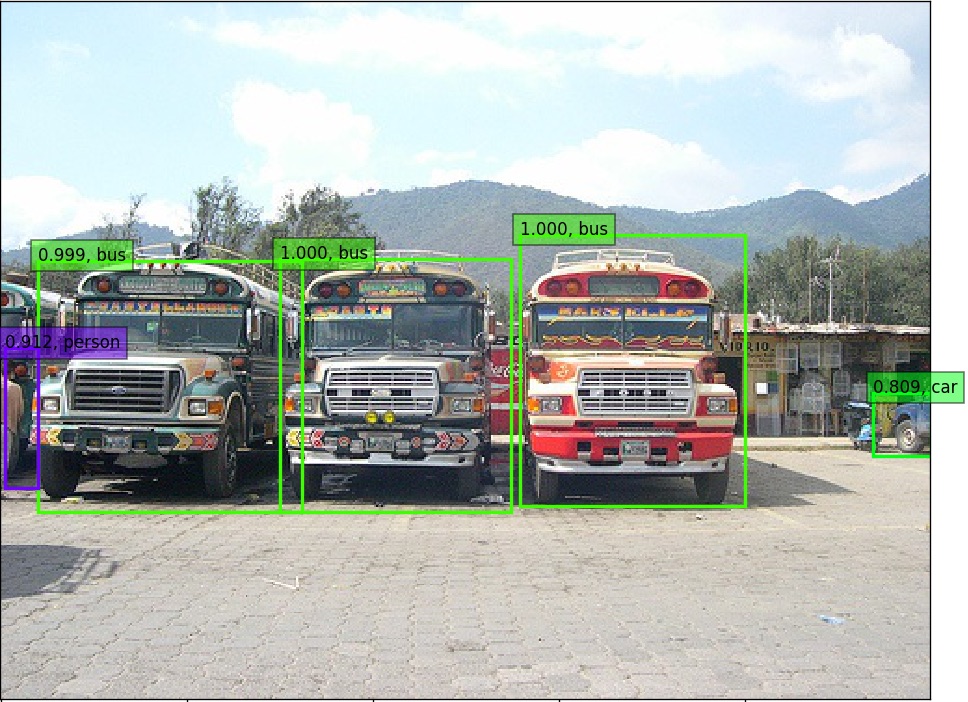}} \\
 \vspace{-2ex}
\subfigure[the {\em sheep} is redundant] {
 \label{O}
 \includegraphics[width=0.46\linewidth]{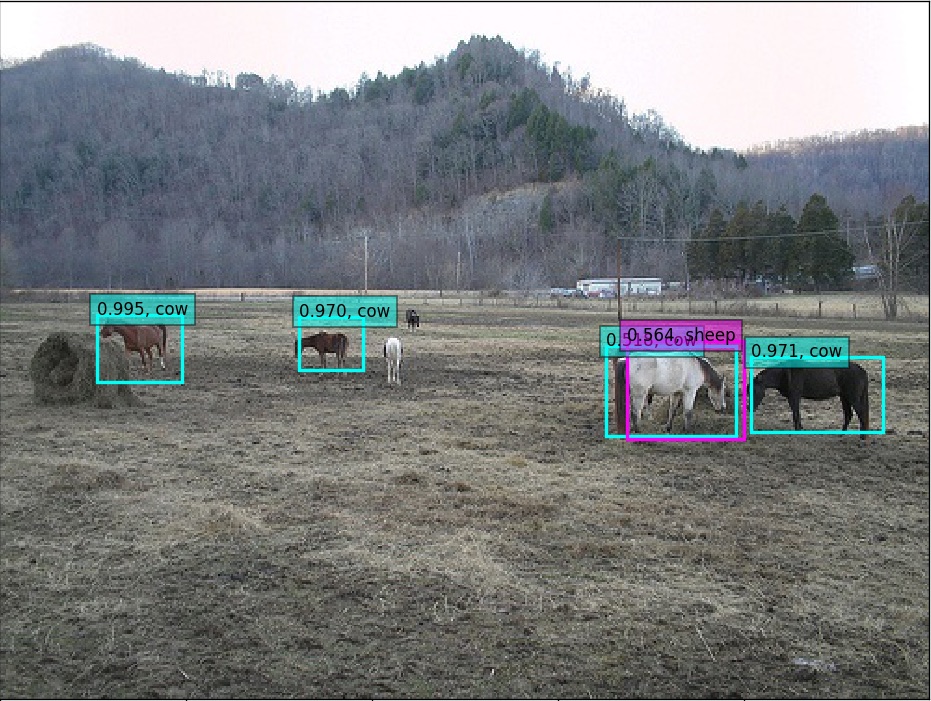}}
  \subfigure[results of {\em Edge}] {
 \label{O}
 \includegraphics[width=0.46\linewidth]{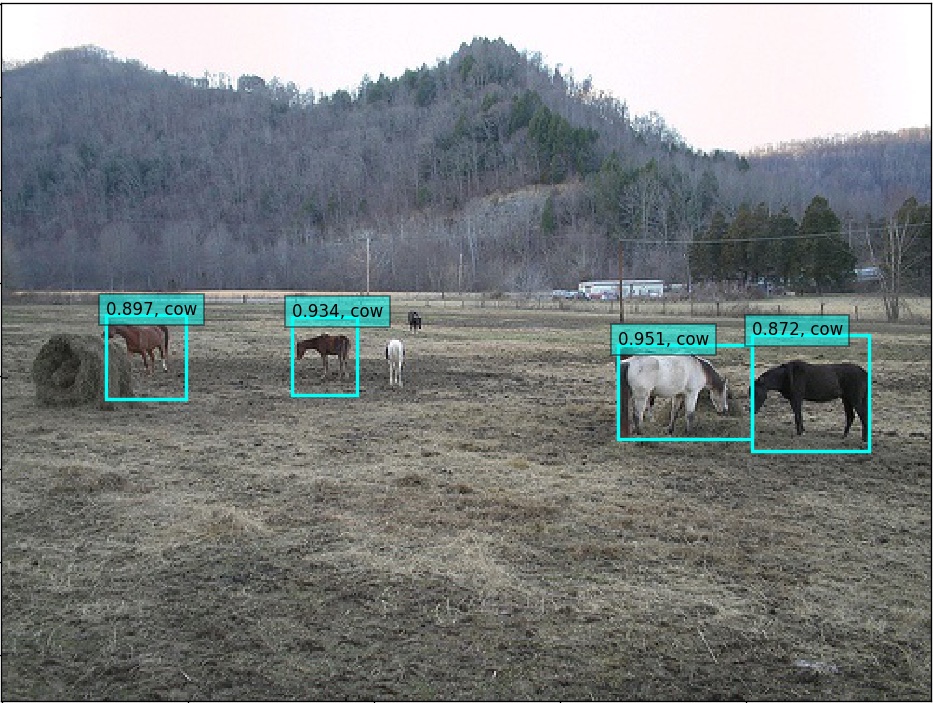}} \\ 
\caption{{\bf Qualitative results of Baseline vs. {\em Edge} on VOC.} In every pair of detection results, the left is based on baseline, and the right is detection result of {\em Edge.}}
\label{fig:edgeRes}
\vspace{-2ex}
\end{figure}

{\bf Relative object visualization.}
As described above in Sec. \ref{sec:SI}, the input of edge GRU is an integrated message from relative nodes for one object. In this part, we check whether the relative object-object relationship has really been learned. For this purpose, we visualize object relationship in an image by edges $e_{j \to i}$. For each node $v_i$, we find the maximum $e_{j \to i}$. If node $i$ and node $j$ are truly detected objects, we draw a dashed line to concatenate box $i$ and $j$ to represent that object $i$ and $j$ have a highly correlated relationship. The results are shown in Fig. \ref{fig:visRela}. 

\begin{figure}[t]
\centering
\subfigure {
 \label{O}
 \includegraphics[width=0.47\linewidth]{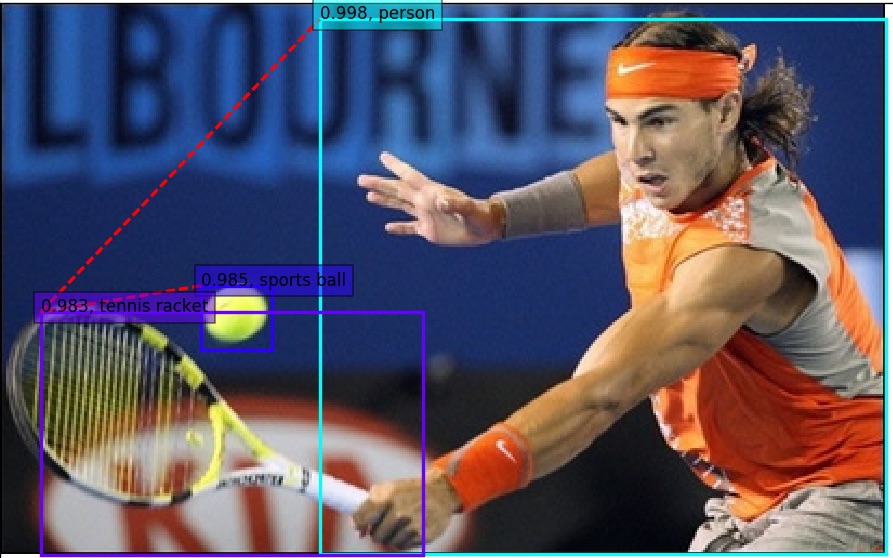}} 
\subfigure{
 \label{O}
 \includegraphics[width=0.44\linewidth]{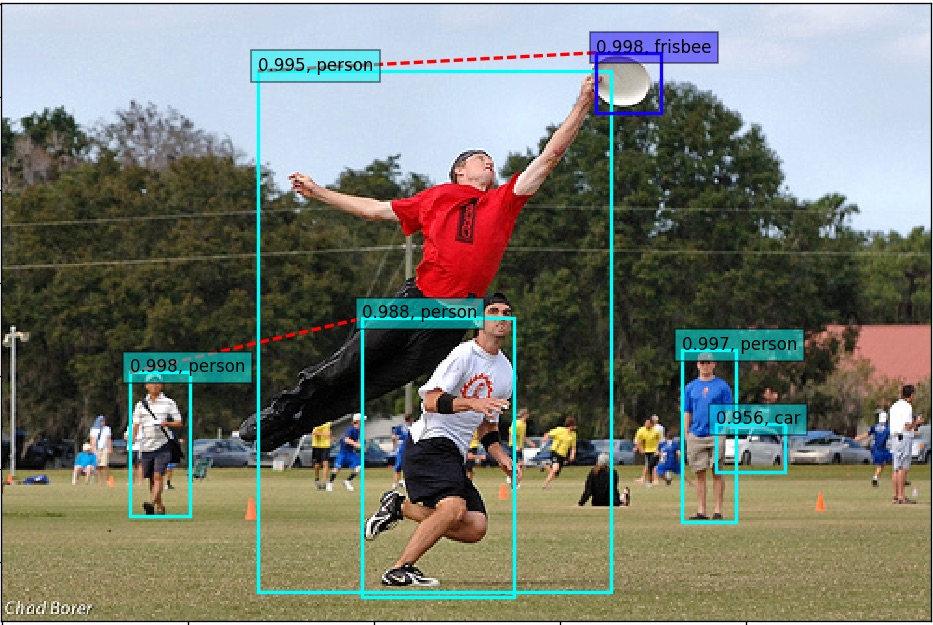}}\\
\caption{{\bf Relative Object Visualization on COCO.} Those objects connected by red dashed line are most relative. {\bf Left}: {\em person} - {\em tennis racket} \& {\em tennis racket} - {\em sports ball}. {\bf Right}: {\em person$^1$} - {\em frisbee} \& {\em person$^2$} - {\em person$^3$}.}
\label{fig:visRela}
\vspace{-1ex}
\end{figure}

\subsection{Ensemble}
\label{sec:way}
At this moment, we have evaluated the effectiveness of two key modules. Then we explore how to conduct an effective fusion of the two separated updated hidden state $h^s$ and $h^e$ of nodes respectively obtained by the modules of {\em Scene} and {\em Edge}.

\begin{table}
\caption{{\bf Performance on VOC 2007 test Using Different Ensemble Ways and Time Steps.} All methods are trained on VOC 07 trainval.}
\begin{center}
{
\begin{tabular}{c|c|c}
\hline
Ensemble Way & Time Steps & mAP \\
\hline
\hline
concatenation & 2 & 70.2 \\
max-pooling & 2 & 70.4 \\
\hline
mean-pooling & 2 & {\bf 70.5} \\
\hline
mean-pooling & 1  & 69.8 \\
mean-pooling & 3 & 69.6 \\
\hline
\end{tabular}}
\end{center}
\label{table:ensemble}
\vspace{-5ex}
\end{table}

{\bf Way of ensemble.}
We explore three ways to integrate these two modules, including max-pooling, mean-pooling and concatenation: $W_a[h^s; h^e]$. From Tab. \ref{table:ensemble}, it can be observed that mean-pooling performs the best.

{\bf Time steps of updating.}
We explore the performance of different numbers of time step. As shown in Tab. \ref{table:ensemble}, our final model achieves the highest performance at training with two time steps, and gradually gets worse afterwards. One possible reason is that
{the graph can form a close loop of message communication after 2 time steps. While with more than 3 time steps, noisy messages start to permeate through the graph.}

\begin{figure}[t]
\vspace{-0.5ex}
\centering
 \includegraphics[width=0.9\linewidth,  trim=45 5 50 30, clip
 ]{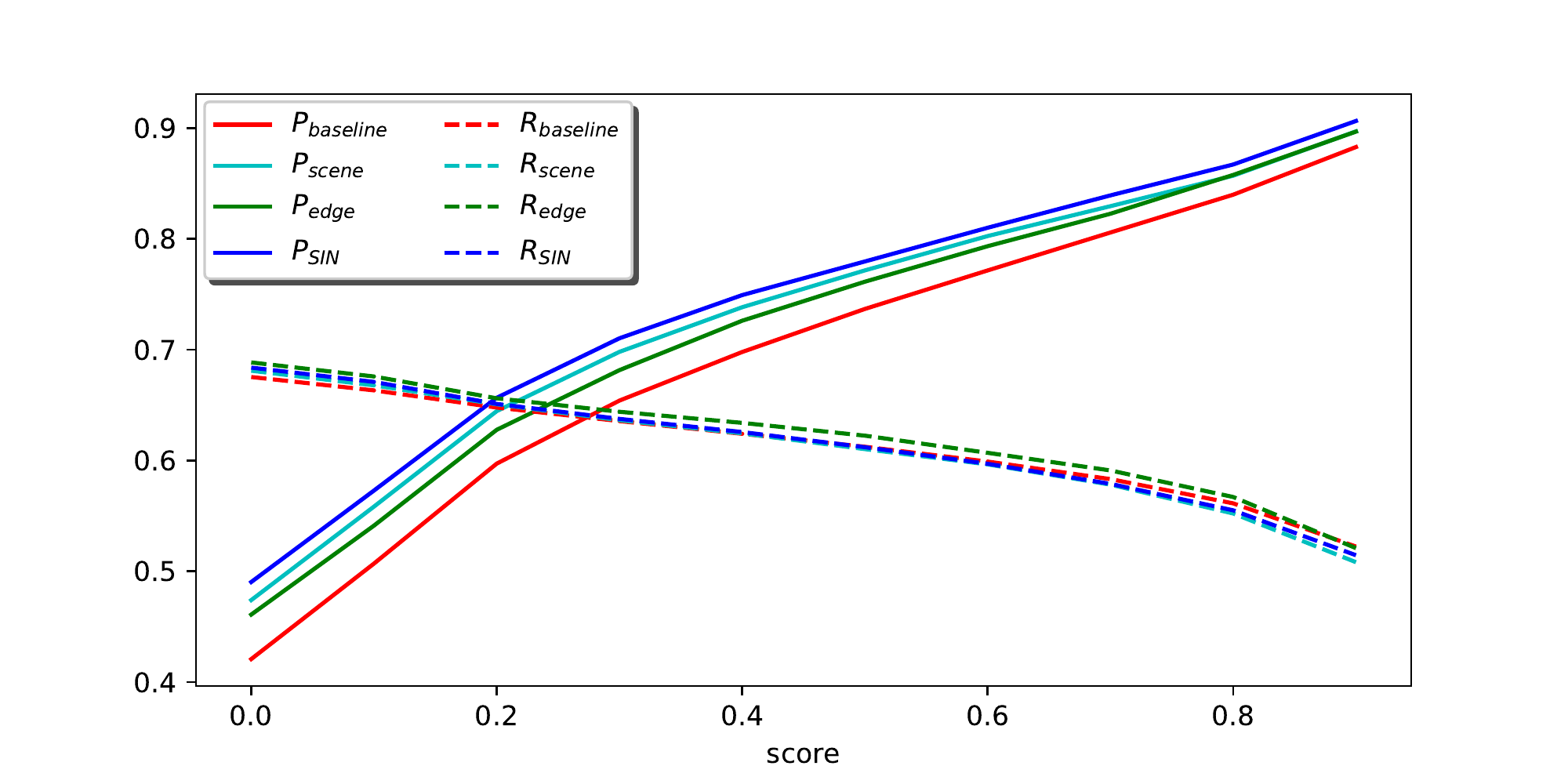}
    \caption{{\bf PR curves.} Legend: {\bf solid line:} precision curve, {\bf dashed line:} recall curve, {\bf red:} baseline. {\bf coral:} {\em Scene}, {\bf green:} {\em Edge}, {\bf blue:} SIN. SIN yields the highest the precision curve, while at the meantime obtains an almost same recall curve compared with the baseline.}
\label{fig:PR}
\vspace{-3ex}
\end{figure}

{\bf Performance of PR curves.}
In this part, we detailedly discuss the performance metrics of our method. At detection score of [0: 0.1: 0.9], we calculate the global precision and recall of detection results by baseline, {\em Scene}, {\em Edge} and SIN ({\em Scene \& Edge}). Then we plot the PR curves in Fig. \ref{fig:PR}. The results show that SIN is able to reach higher precision than the baseline and meanwhile performs almost the same recall, suggesting that when recalling almost the same number of positive instances, our detection results are fewer and more accurate. The limited recall rate might be attributed to the additional relationship constraints which make it more difficult to detect rare samples in a specific scene {\em e.g.} a boat lies on a street. However, detection results using context information are more accurate and confident. This observation exactly manifests the major characteristics of using context information.

\section{Conclusion}
In this paper, we propose a detection method to jointly use scene context and object relationships. In order to effectively leverage these information, we propose a novel structure inference network. Experiments show that scene-level context is important and useful for detection. It particularly performs well on the categories which are highly correlated with scene context, though rare failure cases might happen in case of uncommon situations. As to instance-level relationships, it also plays an important role for object detection, and it could especially improve object localization accuracy. From our current evaluations on VOC to COCO, it is believed that our method has great potential to be applied to larger realistic datasets with more of categories.

\vspace{1ex}
\noindent
{\small
{\bf Acknowledgements.} This work is partially supported by Natural Science Foundation of China under contracts Nos. 61390511, 61772500, 973 Program under contract No. 2015CB351802, CAS Frontier Science Key Research Project No. QYZDJ-SSWJSC009, and Youth Innovation Promotion Association No. 2015085.
}

{\small
\bibliographystyle{ieee}
\bibliography{egbib}
}

\end{document}